\documentclass[letterpaper, 10 pt, conference]{ieeeconf}  % Comment this line out if you need a4paper

\IEEEoverridecommandlockouts                              % This command is only needed if 
\usepackage{graphics} % for pdf, bitmapped graphics files
\graphicspath{ {./Figures/} }
\usepackage{epsfig} % for postscript graphics files
\usepackage{amsmath} % assumes amsmath package installed
\usepackage{amssymb}  % assumes amsmath package installed
\usepackage{hyperref}

\usepackage[ruled,vlined]{algorithm2e}
\usepackage{multirow}

\usepackage{setspace}
\PassOptionsToPackage{hyphens}{url}\usepackage[capitalise]{cleveref}

\title{\LARGE \bf Vehicular Teamwork: Collaborative localization of Autonomous Vehicles}

\author{Jacob Hartzer and Srikanth Saripalli$^{1}$% <-this % stops a space
    \thanks{
        $^{1}$Jacob Hartzer and Srikanth Saripalli are with the Department of Mechanical Engineering, Texas A\&M University, College Station, Texas, USA {\tt\small jmhartzer@tamu.edu}, {\tt\small ssaripalli@tamu.edu}
    }%
}

\usepackage{tikz}
\usepackage{textcomp}
\usepackage{lipsum}

\newcommand\copyrighttext{%
  \footnotesize \textcopyright 2021 IEEE. Personal use of this material is permitted.
  Permission from IEEE must be obtained for all other uses, in any current or future
  media, including reprinting/republishing this material for advertising or promotional
  purposes, creating new collective works, for resale or redistribution to servers or
  lists, or reuse of any copyrighted component of this work in other works.
  DOI: \href{https://doi.org/10.1109/ITSC48978.2021.9564981}{10.1109/ITSC48978.2021.9564981}}
\newcommand\copyrightnotice{%
\begin{tikzpicture}[remember picture,overlay]
\node[anchor=south,yshift=10pt] at (current page.south) {\fbox{\parbox{\dimexpr\textwidth-\fboxsep-\fboxrule\relax}{\copyrighttext}}};
\end{tikzpicture}%
}

\begin{document}

\maketitle
\copyrightnotice
\thispagestyle{empty}
\pagestyle{empty}

%%%%%%%%%%%%%%%%%%%%%%%%%%%%%%%%%%%%%%%%%%%%%%%%%%%%%%%%%%%%%%%%%%%%%%%%%%%%%%%%
\begin{abstract}
    This paper develops a distributed collaborative localization algorithm based on an extended kalman filter. This algorithm incorporates Ultra-Wideband (UWB) measurements for vehicle to vehicle ranging, and shows improvements in localization accuracy where GPS typically falls short.
    The algorithm was first tested in a newly created \href{https://github.com/unmannedlab/collab_localization}{open-source simulation environment} that emulates various numbers of vehicles and sensors while simultaneously testing multiple localization algorithms. Predicted error distributions for various algorithms are quickly producible using the Monte-Carlo method and optimization techniques within MatLab.
    The simulation results were validated experimentally in an outdoor, urban environment. Improvements of localization accuracy over a typical extended kalman filter ranged from 2.9\% to 9.3\% over 180 meter test runs. When GPS was denied, these improvements increased up to 83.3\% over a standard kalman filter.
    In both simulation and experimentally, the DCL algorithm was shown to be a good approximation of a full state filter, while reducing required communication between vehicles.
    These results are promising in showing the efficacy of adding UWB ranging sensors to cars for collaborative and landmark localization, especially in GPS-denied environments. In the future, additional moving vehicles with additional tags will be tested in other challenging GPS denied environments.
\end{abstract}

%%%%%%%%%%%%%%%%%%%%%%%%%%%%%%%%%%%%%%%%%%%%%%%%%%%%%%%%%%%%%%%%%%%%%%%%%%%%%%%%
\section{Introduction}
%%%%%%%%%%%%%%%%%%%%%%%%%%%%%%%%%%%%%%%%%%%%%%%%%%%%%%%%%%%%%%%%%%%%%%%%%%%%%%%%
Currently, there is significant work being performed to develop autonomous vehicles. Such vehicles have the capacity to reduce risk and increase free time of riders. In the development of autonomous vehicles, a key consideration is reliable localization in order to plan for and react to the environment. Typically, high accuracy position is achieved through the use of a Global Navigation Satellite Systems such as GPS. The use of GPS with an Inertial Measurement Unit (IMU) or odometry measurements can lead to an accurate positioning system \cite{Bimbraw}.

Reliable GPS measurements are often not available. Because GPS receivers must have line of sight with at least four satellites, environments that include obstructions pose significant challenges. Urban canyons, tunnels, or indoor environments degrade or fully obstruct GPS signals, forcing a navigation system to rely on dead reckoning \cite{Balamurugan}. These measurements are subject to drift, and therefore cannot be trusted after continued interruption of GPS signal. As such, it would be advantageous to the continued development of autonomous vehicles to use an exteroceptive sensor for navigation through environments where GPS is typically denied.

Ultra-Wideband (UWB), defined as radio technology exceeding 500 MHz or 20\% of the arithmetic center frequency \cite{FCC}, utilizes low-energy pulse communication typically for short-range, high-bandwidth applications. By measuring time of flight across various frequencies, it is possible to measure distance between modules while overcoming multipath errors \cite{Aftanas}. This has allowed UWB modules to be applied towards localization and tracking problems \cite{Location-Aware-UWB}. With the continued development of UWB technology and roll out in commercial products \cite{Apple}, the use of these sensors in navigation systems should continue to be explored.

There are many examples of using UWB ranging modules for high-accuracy indoor localization of ground or aerial systems. These systems typically measure with respect to fixed landmark anchors \cite{Fan,Sun}. There have been examples of utilizing indoor positioning system simulations and physical experiments to predict and show localization accuracy improvements from incorporating UWB measurements to an extended Kalman filter (EKF) \cite{Yao,Benini}. There are also examples of using UWB modules for outdoor vehicular environments. Such a framework has the capacity to increase localization accuracy of existing systems, or provide accurate localization in environments where GPS is not reliable, such as tunnels or urban canyons, as was shown in \cite{Dierenbach}. This has applications in the localization of autonomous vehicles using road landmarks \cite{Martin}.

By using UWB, autonomous vehicles can generate ranging estimates to other independent moving vehicles in a process known as collaborative localization. An example of this is shown in \cite{Ghanem}, where initial experimentation was performed measuring with respect to a fixed vehicle. Given a network of vehicles, it is possible to leverage relative measurements to provide better localization accuracy as a group, than as a collection of individuals. Just like measurements to landmarks, these measurements are not subject to drift when GPS measurements are obscured, and can be made available in both indoor and outdoor environments. These measurements can leverage the growing number of intelligent vehicles on the road, and can increase the accessible area of autonomous vehicles. These collaborative methods can be generally classified in two groups: centralized methods, and decentralized methods.

Centralized collaborative localization (CCL) methods use a single or multiple fusion centers, to which every vehicle communicates measurement information. These state estimators are capable of producing optimal results, and have been used in a number of applications with success \cite{Howard_centralized}. Issues with centralized networks include sensitivity to failure or disconnection, as well as bandwidth and computational power. As the number of measurements made can increase on the order of $\mathcal{O}(n^2)$, centralized networks can meet constraints when large networks of nodes are implemented with complex measurements or update functions.

Decentralized collaborative localization (DCL) methods are defined by distributing the state estimation computation across every agent in the network. Unlike centralized methods, DCL methods are not susceptible to single point failures. These approximations of the centralized method can extend the framework to decrease convergence time \cite{Fabresse} or computational cost without loss of accuracy \cite{Howard_decentralized}. Very promising has been the work in \cite{Luft} which proposed a CL framework that tracks correlations while limiting communication to the two robots that obtain a relative measurement. The algorithm is recursive, reducing memory requirements and it supports generic measurement models, and was implemented using the UTIAS data set, which uses cameras for relative bearing and ranging \cite{UTIAS}.

This paper seeks to expand upon these previous works with an open-sourced simulation framework for testing and validating the DCL algorithm in various scenarios with the Monte-Carlo method. It is with this simulation environment that this paper seeks to validate the efficacy of the DCL algorithm compared to a full state Kalman filter. Additionally, this paper presents experimental results showing improvement of localization accuracy for a vehicle in an urban environment when using the DCL algorithm.

%%%%%%%%%%%%%%%%%%%%%%%%%%%%%%%%%%%%%%%%%%%%%%%%%%%%%%%%%%%%%%%%%%%%%%%%%%%%%%%%
\section{Decentralized Collaborative Localization}
%%%%%%%%%%%%%%%%%%%%%%%%%%%%%%%%%%%%%%%%%%%%%%%%%%%%%%%%%%%%%%%%%%%%%%%%%%%%%%%%
The decentralized collaborative localization algorithm, developed by Luft et. al. \cite{Luft} and reproduced in \cref{alg:prediction,alg:private,alg:relative}, is a form of a Kalman filter that approximates the centralized filter through distributed computation. The estimation of the entire state of a network is broken down into smaller filters where each vehicle has private controls and measurements that are used internally and relative measurements that require the sharing of state and covariance information.

For a network of N vehicles, let $X_i$ be the 6-DOF state of vehicle $i$ such that $X_i = [x,y,\theta, \Dot{x}, \Dot{y}, \Dot{\theta} ]$. Each vehicle has an estimate at time $t$ of its state $\hat{X}^t_i$ and covariance $\Sigma^t_{ii}$. The joint state estimate of the vehicle network is $\hat{X}^t = [\hat{x}^t_1;...;\hat{x}^t_N]$ and the joint covariance is $\Sigma^t = [\Sigma^t_{ij}]_{i,j \in \{1,...,N\}}$. With this network, a decentralized collaborative localization filter can be used.

%%%%%%%%%%%%%%%%%%%%%%%%%%%%%%%%%%%%%%%%%%%%%%%%%%%%%%%%%%%%%%%%%%%%%%%%%%%%%%%%
\subsection{Initialization}
It is assumed upon initialization the vehicle states are uncorrelated. Therefore, the network state can be initialized with the initial estimates $\hat{X}_i$ of each vehicle and the cross correlation terms set to zero.
\begin{equation}
    \{\Sigma^t_{ij} = 0 \}_{i,j \in \{1,...,N\} | i\neq j}
\end{equation}
When the vehicles come into sensing range, generally the cross correlation $\Sigma^{t+1}_{ij} \neq 0$ and therefore the cross-correlation term can be decomposed according to \cite{Bekey}.
\begin{equation}
    \Sigma^{t+1}_{ij} = \sigma^{t+1}_{ij} (\sigma^{t+1}_{ji})^T
\end{equation}
where the simple decomposition $\sigma^{t+1}_{ij} = \Sigma^{t+1}_{ij}$ and $\sigma^{t+1}_{ji} = \mathbb{I}$ is chosen.

%%%%%%%%%%%%%%%%%%%%%%%%%%%%%%%%%%%%%%%%%%%%%%%%%%%%%%%%%%%%%%%%%%%%%%%%%%%%%%%%
\subsection{Prediction}
It is assumed that the vehicles follow the motion model $f$ with process noise $R$ and  control input $u$ which in this case is an IMU measurement. The prediction step for vehicle $i$ is given by the standard EKF equations in \cref{alg:prediction} with the addition of applying the linearized motion model to update the decomposed cross correlation terms $\sigma_{ij}$.

\begin{algorithm}
    \setstretch{1.35}
    \caption{Prediction for Vehicle \textit{i}}
    \label{alg:prediction}
    \KwIn{
    $\hat{X}_i^t$,
    $\Sigma^t_{ii}$,
    $\{ \sigma^t_{ij}\}_{j\in\{1,...,N\}\backslash\{i\}}$,
    $u$}
    \KwOut{
    $\hat{X}_i^{t+1}$,
    $\Sigma^{t+1}_{ii}$,
    $\{ \sigma^{t+1}_{ij}\}_{j\in \{1,...,N\} \backslash\{i\}}$}
    $F = \frac{\partial f (X,u)}{\partial X} (\hat{X}_i^t, u)$ \\
    $\hat{x}_i^{t+1} = f(\hat{x}_i^t, u)$ \\
    $\Sigma^{t+1}_{ii} = F \Sigma^{t}_{ii} F^T + R$ \\
    \For{$ j \in \{1,...,N\} \backslash \{i\} $}{
        $\sigma^{t+1}_{ij} = F \sigma^{t}_{ij}$ \\
    }
\end{algorithm}

%%%%%%%%%%%%%%%%%%%%%%%%%%%%%%%%%%%%%%%%%%%%%%%%%%%%%%%%%%%%%%%%%%%%%%%%%%%%%%%%
\subsection{Private Update}

It is assumed that the private update measurements are functions of the state of a single vehicle with a Gaussian error disturbance.
\begin{equation}
    z^t_i = h({\hat{x}}^{t}_i) + \nu_p
\end{equation}
with $\nu_p \sim \mathcal{N}(0,R^t_i)$, $R^t_i$ being the measurement noise, and $g$ being the measurement model. The private update step for the system is shown in \cref{alg:private}. Again, the algorithm is very similar to the typical update step of an EKF with the addition of updates to the decomposed cross correlation terms.

\begin{algorithm}
    \setstretch{1.35}
    \caption{Private Update for Vehicle \textit{i}}
    \label{alg:private}
    \KwIn{
    $\hat{X}_i^t$,
    $\Sigma^t_{ii}$,
    $\{ \sigma^t_{ij}\}_{j\in \{1,...,N\} \backslash\{i\}}$,
    $z$}
    \KwOut{
    $\hat{X}_i^{t+1}$,
    $\Sigma^{t+1}_{ii}$,
    $\{ \sigma^{t+1}_{ij}\}_{j\in \{1,...,N\} \backslash\{i\}}$}
    $H = \frac{\partial h (X)}{\partial X} (\hat{X}_i^t)$ \\
    $K_i = \Sigma^t_{ii} H^T (H \Sigma^t_{ii} H^T + Q)^{-1}$ \\
    $\hat{X}_i^{t+1} = \hat{X}^t_i + K_i [z-h(\hat{X}^t_i)]$ \\
    $\Sigma^{t+1}_{ii} = (\mathbb{I} - K_i H) \Sigma^{t}_{ii}$ \\
    \For{$ j \in \{1,...,N\}\backslash \{i\}$}{
        $\sigma^{t+1}_{ij} =  (\mathbb{I} - K_i H) \sigma^{t}_{ij}$
    }
\end{algorithm}

%%%%%%%%%%%%%%%%%%%%%%%%%%%%%%%%%%%%%%%%%%%%%%%%%%%%%%%%%%%%%%%%%%%%%%%%%%%%%%%%
\subsection{Relative Update}

Lastly, it is assumed that the relative update measurement is a function of the state of two vehicles $i$ and $j$, with a Gaussian error disturbance.
\begin{equation}
    z^t_i = g(\hat{x}^{t}_i, \hat{x}^{t}_j) + \nu_p
\end{equation}
with $\nu_p \sim \mathcal{N}(0,R^t_i)$, $R^t_i$ being the measurement noise, and $g$ being the measurement model. These updates come from vehicle-to-vehicle relative UWB ranging measurements. The relative update step for the system is shown in \cref{alg:relative}. The update contains updates to the state estimates and covariances of the two vehicles involved in the relative measurement by combining the decomposed cross correlation terms. As explained in \cite{Luft}, the inability to simplify the cross correlation update equation for all non-involved vehicles $k$ into the form
\begin{equation}
    \sigma^{t+1}_{ab} = A \sigma^{t}_{ab} ~|~a \in \{i,j\}, b\in \{1,...,N\}\backslash\{i,j\}
\end{equation}
creates difficulty when trying to minimize necessary communication between vehicles. The update equation used is not only stable, but also also a very good approximation, as will be shown in simulation.

\begin{algorithm}
    \setstretch{1.35}
    \caption{Relative Update for Vehicles \textit{i,j}}
    \label{alg:relative}
    \KwIn{
    $\hat{X}_i^t$,
    $\Sigma^t_{ii}$,
    $\{ \sigma^t_{ij}\}_{j\in\{1,...,N\}\backslash\{i\}}$,
    $r$}
    \KwOut{
    $\hat{X}_i^{t+1}$,
    $\Sigma^{t+1}_{ii}$,
    $\{ \sigma^{t+1}_{ij}\}_{j\in\{1,...,N\}\backslash\{i\}}$}
    \If{
    vehicle $i$ detects vehicle $j$}{
    receive from vehicle $j$: $\hat{X}^t_j$, $\Sigma^t_{jj}$, $\sigma^t_{ji}$ \\
    $\Sigma^t_{ij} = \sigma^t_{ij}(\sigma^t_{ji})^T  $ \\
    $\Sigma^t_{aa} = \begin{bmatrix}
            \Sigma^t_{ii} & \Sigma^t_{ij} \\ (\Sigma^t_{ij})^T & \Sigma^t_{jj}
        \end{bmatrix}$ \\
    $H_a =
        \begin{bmatrix}
            \frac{\partial h(X_i,X_j)}{\partial X_i}(\hat{X}^t_i, \hat{X}^t_i), &
            \frac{\partial h(X_i,X_j)}{\partial X_i}(\hat{X}^t_i, \hat{X}^t_i)
        \end{bmatrix}$ \\
    $K_a = \Sigma^t_{aa} H^T_a(H_a\Sigma^t_{aa}H^T_{a} + Q)^{-1}$ \\
    $\begin{bmatrix}
            \hat{X}^{t+1}_i \\ \hat{X}^{t+1}_j
        \end{bmatrix} =
        \begin{bmatrix}
            \hat{X}^{t}_i \\ \hat{X}^{t}_j
        \end{bmatrix} + K_a
        \begin{bmatrix}
            r - h(\hat{X}^t_i, \hat{X}^t_i)
        \end{bmatrix}$\\
    $\begin{bmatrix}
            \Sigma^t_{ii} & \Sigma^t_{ij} \\ (\Sigma^t_{ij})^T & \Sigma^t_{jj}
        \end{bmatrix} = (\mathbb{I} - K_a H) \Sigma^t_{aa} $\\
    send to vehicle $j$: $\hat{X}^{t+1}_j$, $\Sigma^{t+1}_{jj}$ \\
    $\sigma^{t+1}_{ij} = \Sigma^{t+1}_{ij}$\\
    \For{$k \in \{1,...,N\} \backslash \{i,j\} $}{
        $\sigma^{t+1}_{ik} = \Sigma^{t+1}_{ii}(\Sigma^{t}_{ii})^{-1} \sigma^{t}_{ik}$}}
    \If{vehicle $j$ is detected by vehicle $i$}{
    send to vehicle $i$: $\hat{X}^t_j$, $\Sigma^t_{jj}$, $\sigma^t_{ji}$ \\
    receive from vehicle $i$:
    $\hat{X}^{t+1}_j$, $\Sigma^{t+1}_{jj}$ \\
    $\sigma^{t+1}_{ij} = \mathbb{I}$\\
    \For{$k \in \{1,...,N\} \backslash \{i,j\} $}{
        $\sigma^{t+1}_{jk} = \Sigma^{t+1}_{jj}(\Sigma^{t}_{jj})^{-1} \sigma^{t}_{jk}$}}
\end{algorithm}

%%%%%%%%%%%%%%%%%%%%%%%%%%%%%%%%%%%%%%%%%%%%%%%%%%%%%%%%%%%%%%%%%%%%%%%%%%%%%%%%
\section{Simulation Design}
%%%%%%%%%%%%%%%%%%%%%%%%%%%%%%%%%%%%%%%%%%%%%%%%%%%%%%%%%%%%%%%%%%%%%%%%%%%%%%%%
In order to simulate the efficacy of the collaborative localization algorithm, an \href{https://github.com/unmannedlab/collab_localization}{open-source simulation framework} that would be flexible in the number of cars and their positions was created in MatLab \cite{collab_loc}. This simulation included sensing models for IMU, wheel odometry, GPS, and UWB measurements between both stationary and moving landmarks. The simulation was vectorized and parallelized to greatly increase processing speed for large Monte-Carlo simulations, and allows for error distributions to be quickly estimated given any vehicle scenario. Lastly, multiple localization algorithms can be run in parallel to directly compare performance of each with the same set of random sensor measurements. This, combined with the ability to simultaneous simulation thousands of experiments, allows for direct comparisons of error distributions.

%%%%%%%%%%%%%%%%%%%%%%%%%%%%%%%%%%%%%%%%%%%%%%%%%%%%%%%%%%%%%%%%%%%%%%%%%%%%%%%%
\subsection{Sensing Models}
The sensing error models used in the simulation all took the form of zero-mean Gaussian distributions.
\begin{equation}
    z = h({\hat{x}}) + \nu
\end{equation}
With there exist better models for the various sensors used, this assumption allows for sufficient testing of the various algorithms, without sacrificing speed due to computational complexity. Specifically, there is significant computational advantage in generating large vectors of normally distributed random numbers, and this advantage is what was leveraged to make the Monte-Carlo simulation run on the order of minutes rather than hours.

\subsection{Simulation Output}
The resulting output of this software package is the RMS error of position and heading for each simulated vehicle for every simulation run. The number of simulations run can range from a single evaluation to up to 100,000 runs, depending on system memory. These results can be plot as a distribution of errors for comparing various parallel localization algorithms. When compiled in such Monte Carlo simulations, these data can be used to evaluate filter performance improvements.

\begin{figure}
    \centering
    \includegraphics[width = 0.8\linewidth]{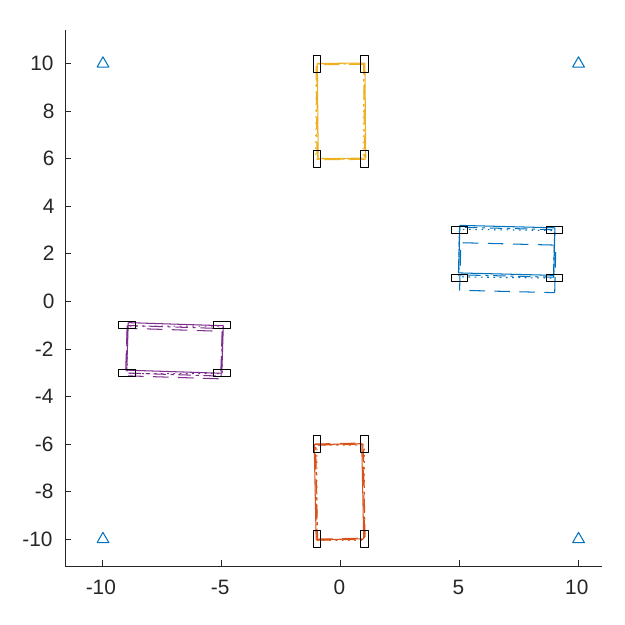}
    \caption{Street crossing example of using the simulation with multiple parallel localization algorithms}
    \label{fig:sim}
\end{figure}

\section{Simulation Results}

Using the simulation framework, it was possible to test a wide variety of vehicle environments and sensor configurations. These configurations, with an example shown in \cref{fig:sim}, included vehicles moving parallel to each other, vehicles moving perpendicular in street crossings, and moving in groups in tunnels without GPS measurements but with UWB landmarks. Additionally, it was possible to add stationary UWB landmarks to each of these situations.

In each of the three different scenarios, three different localization algorithms were tested: Extended Kalman Filter (EKF), Centralized Collaborative Localization (CKF), and Decentralized Collaborative Localization (DCL). Note that the CCL algorithm is a full-state EKF that receives every sensor message and maintains a single state estimate for all vehicles. The CCL algorithm represents the theoretical maximal performance of a DCL algorithm, and is therefore a good test of the assumptions made and check for overconfidence.

\begin{figure}
    \centering
    \includegraphics[width = 0.8\linewidth]{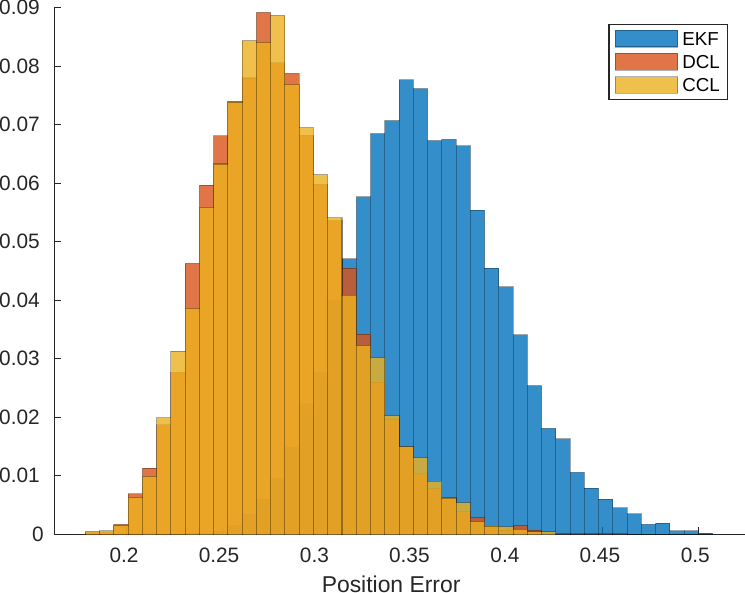}
    \caption{Monte Carlo Simulation with 10,000 runs of two  collaborative vehicles moving in parallel}
    \label{fig:sim_par}
\end{figure}

\begin{figure}
    \centering
    \includegraphics[width = 0.8\linewidth]{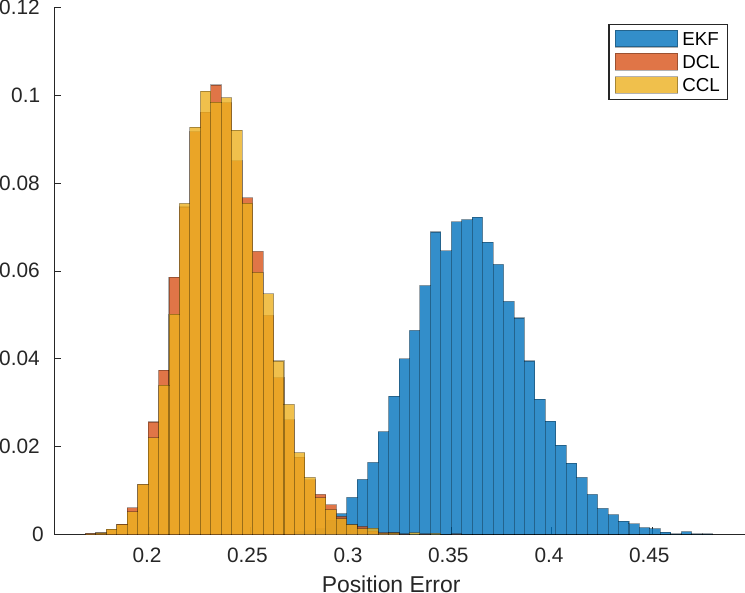}
    \caption{Monte Carlo Simulation with 10,000 runs of two collaborative vehicles moving in parallel with landmarks every 50 meters}
    \label{fig:sim_scs}
\end{figure}

\begin{figure}
    \centering
    \includegraphics[width = 0.8\linewidth]{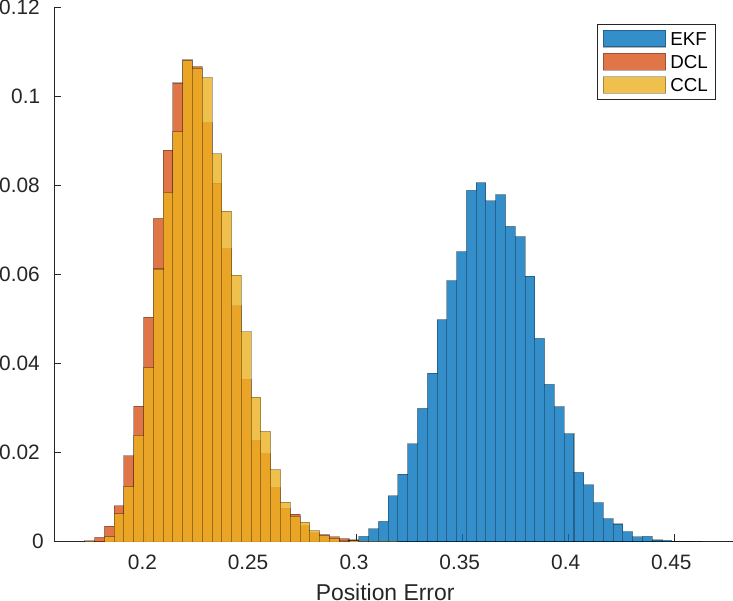}
    \caption{Monte Carlo Simulation with 10,000 runs of two collaborative vehicles moving in parallel with landmarks every 50 meters}
    \label{fig:sim_tun}
\end{figure}

The resulting error distributions from the Monte Carlo simulations performed are shown in \cref{fig:sim_par,fig:sim_scs,fig:sim_tun}. These figures show position localization accuracy of three algorithms run in parallel in each of the simulated scenarios. A numerical summary of the simulation results is shown in \cref{tab:sim_results}.

These simulations showed that collaborative UWB measurements offered improvements to localization accuracy across all experiments. As would be expected, using more vehicles improved accuracy, and using stationary landmarks also offered significant improvements. This was especially the case in situations where GPS data was not available such as in a tunnel environment.

Additionally, it should be noted the CCL and DCL algorithms performed, on average, within 1 cm of accuracy between each other across the simulation scenarios. While this research originally sought to improve the update equation approximation made in the DCL algorithm, this result helps to validate this method for very closely approximate a centralized Kalman Filter without requiring large amounts of communication between vehicles. Additionally, these simulations also validates the update equation approximation does not result in overconfidence of the filter, as divergence was not see in any of the hundreds of thousands of simulated experiments.

Therefore, the simulations validated the DCL algorithm for it's use in collaborative localization of autonomous vehicles and that the approximation to decentralize the filter did not induce a noticeable reduction in accuracy.

\begin{table}
    \centering
    \begin{tabular}{c|c|c|c|c|c}
        Simulation & GPS & EKF  & DCL  & CCL  \\ \hline
        Parallel   & Yes & 0.19 & 0.15 & 0.15 \\
        Crossing   & Yes & 0.19 & 0.13 & 0.13 \\
        Tunnel     & No  & 0.19 & 0.12 & 0.12
    \end{tabular}
    \caption{Summary of Monte-Carlo simulation RMS position errors, in meters}
    \label{tab:sim_results}
\end{table}

%%%%%%%%%%%%%%%%%%%%%%%%%%%%%%%%%%%%%%%%%%%%%%%%%%%%%%%%%%%%%%%%%%%%%%%%%%%%%%%%
\section{Experimental Results}
%%%%%%%%%%%%%%%%%%%%%%%%%%%%%%%%%%%%%%%%%%%%%%%%%%%%%%%%%%%%%%%%%%%%%%%%%%%%%%%%

\begin{figure}
    \centering
    \includegraphics[width = \linewidth]{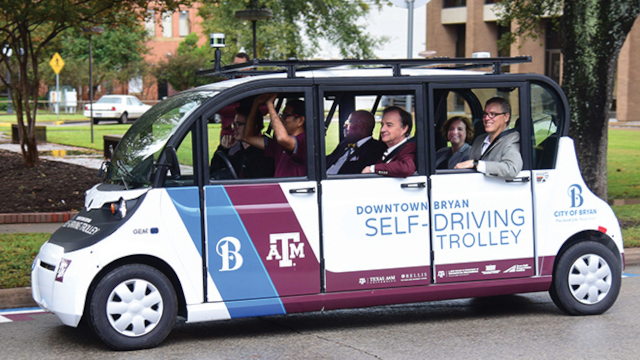}
    \caption{Unmanned Systems Lab Autonomous Trolley}
    \label{fig:trolley}
\end{figure}

The filter framework was tested in downtown Bryan, Texas using the Unmanned Systems Lab autonomous trolley shown in  \cref{fig:trolley}. The Decwave EVK1000 UWB ranging evaluation boards were mounted alongside a VectorNav VN-300 INS. The trolley itself includes PACMod, a by-wire kit prepared by Autonomous Stuff, which gives access to wheel odometry and steering angle measurements. These sensors formed the basis of the filters, while a secondary real-time kinematic GPS (RTK GPS) system from ArduSimple was used as ground truth for filter performance evaluation as it can obtain accuracy to 2 cm.

Using tripods, additional UWB ranging modules were placed in various experimental setups to either represent stationary landmarks or non-stationary vehicles. Due to limitations of in-person testing, only one moving vehicle was used and other tags were simulated as vehicles using a steady state covariance derived from the trolley at rest for a long period of time. Due to the nature of the UWB modules making available all measurements, a vehicle-to-vehicle network was not necessary in order to run any of the three algorithms on a single trolley's computer. The Robot Operating System (\href{https://www.ros.org/}{ROS}) middleware was used to make available the sensor measurements as well as run the separate localization algorithms.

Using this setup, the simulated environment was recreated experimentally in the three different configurations listed and RMS position errors over the course of 180 meter runs were calculated with respect to the on board RTK GPS.

As in the simulations, the addition of the UWB sensors on the vehicle offered improvements to the localization accuracy of the vehicle in all scenarios. This improvement was even more pronounced when GPS data was not available in the tunnel scenario. A summary of the results are shown in \cref{tab:exp_sum}. The DCL algorithm, again, performed as well as the CCL algorithm, down to 1 cm. This further validates the approximations made led to accurate yet not overconfident updates to cross correlation covariance matrices.

Interestingly, despite using the same intrinsic sensor errors in simulation, the experimental errors were significantly higher. It is believed that this occurred due to poor GPS signal in the testing environment. Despite this, relative improvements were still seen due to the addition of UWB measurements and collaborative localization. These improvements, as expected, were lower when GPS was available, producing improvements of 2.9\% and 9.3\% for the Parallel and Crossing scenarios respectively. However, without GPS, the system was subject to drift and therefore saw significant improvements when adding UWB: 83.3\% improvement in the tunnel scenario.

Therefore, in all cases, experimentation showed the addition of UWB ranging to other vehicles led to improvements in localization accuracy, with the most significant improvements seen when GPS was unavailable.

\begin{table}[ht]
    \centering
    \begin{tabular}{c|c|c|c|c}
        Experiment & GPS & EKF  & DCL  & CEKF \\ \hline
        Parallel   & Yes & 2.06 & 1.81 & 1.81 \\
        Crossing   & Yes & 2.32 & 1.61 & 1.61 \\
        Tunnel     & No  & 7.78 & 1.30 & 1.30
    \end{tabular}
    \caption{Summary of experimental RMS position errors, in meters, over a distance of 180 meters}
    \label{tab:exp_sum}
\end{table}

\begin{figure}
    \centering
    \includegraphics[width = 0.8\linewidth]{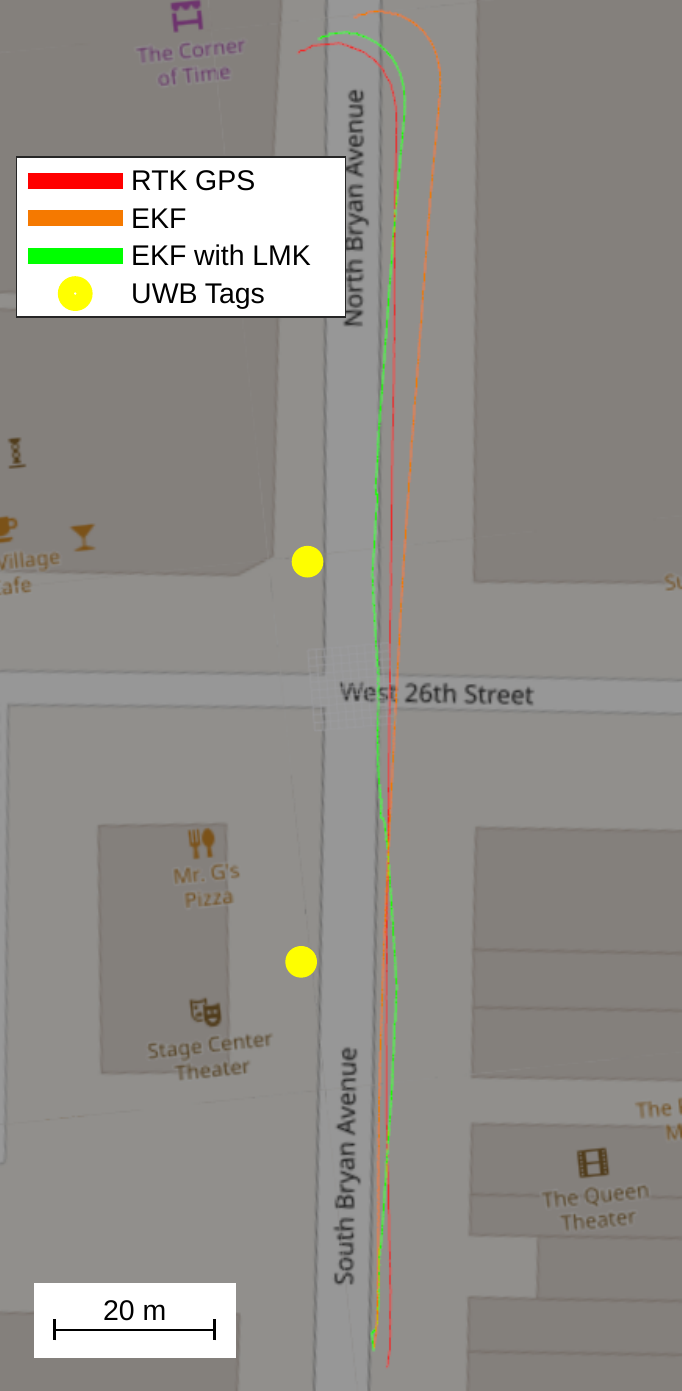}
    \caption{Experimental results of DCL algorithm in tunnel scenario}
    \label{fig:exp_tun}
\end{figure}

%%%%%%%%%%%%%%%%%%%%%%%%%%%%%%%%%%%%%%%%%%%%%%%%%%%%%%%%%%%%%%%%%%%%%%%%%%%%%%%%
\section{Future Work}
%%%%%%%%%%%%%%%%%%%%%%%%%%%%%%%%%%%%%%%%%%%%%%%%%%%%%%%%%%%%%%%%%%%%%%%%%%%%%%%%
In the future, testing will include more in-person experiments with multiple moving vehicles and a greater number of UWB tags. This would allow for better evaluation of truly cooperative localization of vehicles using UWB and will further validate the algorithm used in simulation. This will also explore the use of real-time varying covariance estimations in the update equations and any issues with measurement delays and networking between vehicles. Additionally, transitioning into GPS-denied environments will be explored such as vehicles entering tunnels or warehouses, and how the use of stationary UWB landmarks can aid in navigating these challenging environments.

%%%%%%%%%%%%%%%%%%%%%%%%%%%%%%%%%%%%%%%%%%%%%%%%%%%%%%%%%%%%%%%%%%%%%%%%%%%%%%%%
\section{Conclusions}
%%%%%%%%%%%%%%%%%%%%%%%%%%%%%%%%%%%%%%%%%%%%%%%%%%%%%%%%%%%%%%%%%%%%%%%%%%%%%%%%
In conclusion, this paper presented the creation of an open-source collaborative localization simulation package as well simulated and experimental results of collaborative localization of autonomous vehicles. The simulation is capable of handing user defined scenarios with multiple vehicles, and can quickly create error distributions using the Monte-Carlo method and optimization techniques within MatLab.

In both simulations and in experiments, this work validated the DCL algorithm is a very good approximation of a full state CCL algorithm, while significantly reducing the amount of communication required between vehicles. Additionally, it was shown in simulation and experimentally the inclusion of collaborative localization improved the localization accuracy of autonomous vehicles, especially when in environments without GPS measurements available. Improvements in localization accuracy of 2.9\% and 9.3\% were shown when vehicles were moving in parallel and through an intersection, respectively. While vehicles were moving in a simulated tunnel environment without GPS, the improvement due to collaboration was 83.3\%.

Therefore, methods of collaborative localization show promise for increasing localization accuracy of autonomous vehicles, especially in applications where GPS data is not readily available. UWB ranging is a viable way to include these types of measurements, and its application will continue to be explored with increased numbers of measurements as well as applications to other challenging environments.

% \addtolength{\textheight}{-12cm}   

%%%%%%%%%%%%%%%%%%%%%%%%%%%%%%%%%%%%%%%%%%%%%%%%%%%%%%%%%%%%%%%%%%%%%%%%%%%%%%%%

%%%%%%%%%%%%%%%%%%%%%%%%%%%%%%%%%%%%%%%%%%%%%%%%%%%%%%%%%%%%%%%%%%%%%%%%%%%%%%%%



% \section*{Disclaimer}

% Lorem ipsum dolor sit amet, consectetur adipiscing elit, sed do eiusmod tempor incididunt ut labore et dolore magna aliqua. Ut enim ad minim veniam, quis nostrud exercitation ullamco laboris nisi ut aliquip ex ea commodo consequat. Duis aute irure dolor in reprehenderit in voluptate velit esse cillum dolore eu fugiat nulla pariatur. Excepteur sint occaecat cupidatat non proident, sunt in culpa qui officia deserunt mollit anim id est laborum.

% \section*{Acknowledgment}

% Support for this research was provided in part by a grant from the North Texas Tollway Authority. 

%%%%%%%%%%%%%%%%%%%%%%%%%%%%%%%%%%%%%%%%%%%%%%%%%%%%%%%%%%%%%%%%%%%%%%%%%%%%%%%%

\bibliographystyle{IEEEtran}
\bibliography{IEEEabrv,references}

\end{document}